\documentclass[journal]{IEEEtran}
\usepackage[cmex10]{amsmath}
\usepackage{bm}
\usepackage[switch]{lineno}
\usepackage{caption}
\usepackage{amsfonts}
\usepackage{booktabs}
\usepackage{comment}
\usepackage{subfigure}
\usepackage{graphicx}
\usepackage{epstopdf}
\usepackage{multirow}
\usepackage{verbatim}

\usepackage{color}
\ifCLASSINFOpdf
\else
\fi
\begin{document}
\title{AttKGCN: Attribute Knowledge Graph Convolutional Network for Person Re-identification}

\author{Bo~Jiang,
        Xixi~Wang
        and~Jin~Tang% <-this % stops a space
\IEEEcompsocitemizethanks{\IEEEcompsocthanksitem Bo Jiang, Xixi Wang and Jin Tang (corresponding author) are Anhui University, Hefei, China.\protect\\
E-mail: jiangbo@ahu.edu.cn}}
\maketitle

\begin{abstract}
Discriminative feature representation of person image is important for person re-identification (Re-ID) task.
Recently, attributes have been demonstrated beneficially in guiding for learning more discriminative feature representations for Re-ID.
As attributes normally co-occur in person images, it is desirable
to model the attribute dependencies to improve the attribute prediction and thus Re-ID results.
In this paper, we propose to model these attribute dependencies
via a novel attribute knowledge graph (AttKG), and propose
 a novel Attribute Knowledge Graph Convolutional Network (AttKGCN) to solve Re-ID problem.
 AttKGCN integrates both attribute prediction and Re-ID learning together in a unified end-to-end framework which can boost their performances, respectively.
AttKGCN first builds a directed attribute KG whose nodes denote attributes and edges
encode the co-occurrence relationships of different attributes.
Then, AttKGCN learns a set of inter-dependent attribute classifiers which are combined with person visual descriptors for attribute prediction.
Finally, AttKGCN integrates attribute description and deeply visual representation together to construct a more discriminative feature representation for Re-ID task.
Extensive experiments on several benchmark datasets demonstrate the effectiveness of AttKGCN on attribute prediction and Re-ID tasks.
\end{abstract}

%%%%%%%%% BODY TEXT
\section{Introduction}

Object (\emph{e.g.}, person, vehicle) re-identification (Re-ID) is an active research problem in computer vision area. % The aim of Re-ID tasks is to re-identify a query object image from a set of object images taken by multiple cameras.
%~\cite{tay2019aanet,sun2019perceive,si2018dual,Sun2017SVDNet,shen2018deep,chen2018group,li2018harmonious,Chen_2018_CVPR,ustinova2017multi}.
Many of existing Re-ID methods adopt a classification framework which aims to determine the label of an input person image by using a classifier trained on the training samples~\cite{su2017pose,si2018dual,chen2018group,sun2018beyond,sun2019perceive,shen2018person,yan2019learning}.
% The classifier is trained in the first frame by using the ground truth bounding box, and then updated in the subsequent frames based on the current tracked result.
%  in which the foreground object trackers are often disturbed by the introduced background information. To alleviate the influences of background information,
Although recent years have witnessed rapid advancements in Re-ID, it is still a challenging task due to large changes of object visual appearance caused by  pose, illumination, deformation and occlusion, etc.

%example
\begin{figure}[ht]
\centering
\includegraphics[width=0.4\textwidth]{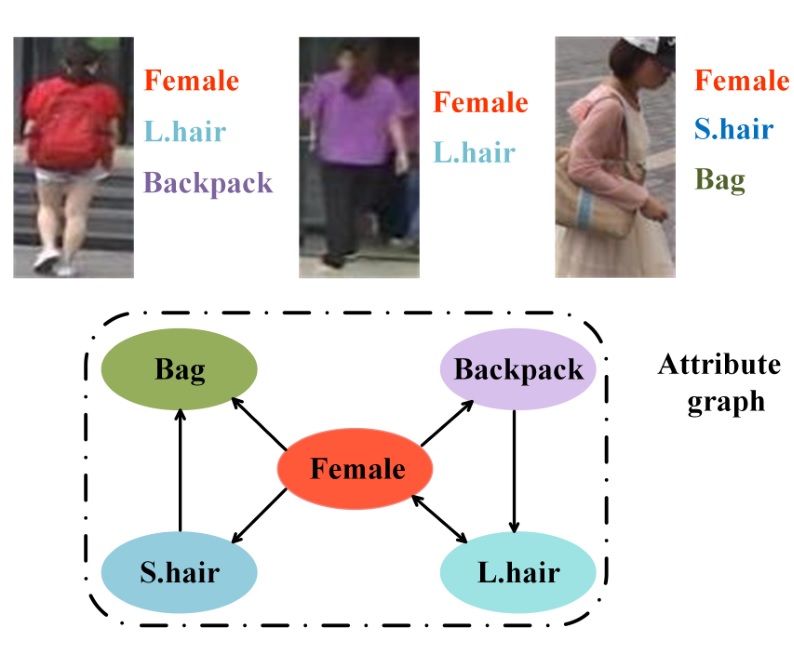}
\caption{An example of attribute knowledge graph $G(A, P)$. Each node denotes a specific attribute $A_i$, such as Bag, Female, etc. A direct edge exists between each attribute pair $\{A_i, A_j\}$ which encodes the  co-occurrence relationship $P_{ij}$ between them. The detail computation of $P_{ij}$ is presented in section~\ref{sec:3.2}. }
\label{fig:example}
\end{figure}

One main issue for Re-ID problem is how to generate strong discriminative feature representation for person images~\cite{matsukawa2016person,chang2018multi,lin2019improving}.
% Deeply-learned features have been verified stronger discriminative ability for person Re-ID~\cite{su2016deep,chang2018multi,lin2019improving}. %, especially when aggregated from deeply-learned attribute  features
Recently, attributes have been aggregated with deeply-learned Re-ID model and demonstrated beneficially in guiding Re-ID model  to learn a stronger discriminative feature representation which can obviously improve the final Re-ID results~\cite{schumann2017person, han2018attribute,ijcai2018-153}.
%In particular, deeply-learned features have been verified stronger discriminative ability, especially when aggregated from deeply-learned attribute  features~\cite{su2016deep,chang2018multi,lin2019improving,tay2019aanet}.
For example,
%[\textbf{please list some deeply learned attribute base reid}]
%\cite{su2016deep} proposed a semi-supervised attribute learning architecture, which is pre-trained on the source datasets with attributes label and IDs label, respectively, for the sake of learning certain discrimination ability to predict attribute labels for the target dataset.
Schumann $et\ al.$~\cite{schumann2017person} propose to first train an attribute-classifier and then incorporate it into person Re-ID model.
Chang $et\ al.$~\cite{chang2018multi} develop Multi-Level Factorisation Net (MLFN) to learn
some latent discriminative factors for each person image based on multiple semantic levels.
Li $et\ al.$~\cite{li2019attributes} propose Attributes-aided Part detection and Refinement network (APDR)
 which aims to employ the attribute learning to handle the body part misalignment and thus local feature extraction. % employing attribute learning. %exploiting
%In order to solve the part misalignment, they proposed Attributes-aided Part detection and Refinement network(APDR) to identify people by leveraging the attribute learning process as a part localizer.
Tay $et\ al.$~\cite{tay2019aanet} propose Attribute Attention Network (AANet) which
integrates  person attributes and attribute attentions together in a unified classification framework for Re-ID problem.

As attributes normally co-occur in person images, it is desirable
to model the attribute dependencies to improve the attribute prediction and thus final Re-ID results.
However, existing works generally fail to exploit them for Re-ID problem.
To capture and explore such important dependencies, in this paper, we first build an \emph{Attribute Knowledge Graph} (AttKG) whose  nodes denote attributes and edges encode the co-occurrence relationships among different attributes, as shown in Figure~\ref{fig:example}.
Then, inspired by recent works~\cite{wang2018zero,chen2019multi}, we propose a novel Attribute Knowledge Graph Convolutional Network (AttKGCN) which integrates both attribute prediction and Re-ID learning together in a unified end-to-end classification framework.
% which can boost their respectively performances.
%AttKGCN first builds a directed attribute KG whose nodes denote attributes and edges
%encode the co-occurrence relationships of different attributes.
AttKGCN can learn a set of inter-dependent attribute classifiers which are combined with each person visual descriptor for its attribute prediction.
%Finally, AttKGCN integrates attribute description and deeply visual representation together to construct a more discriminative feature representation for Re-ID task.
%
%we propose a novel Attribute Graph Convolutional Network (AttGCN) for person attribute prediction and re-identification.
%The proposed AttGCN first builds a directed attribute graph whose nodes denote attributes and edges represent the co-occurrence relationship among different attributes, as shown in Figure 1.
%Then, AttGCN learns a set of inter-dependent attribute classifiers via GCN based mapping function. These classifiers are applied to the person visual descriptors for the attribute prediction.
%Finally, AttGCN integrates attribute description and deeply visual representation together to construct a more discriminative feature representation for the final re-identification task.

Overall, the main contributions of this paper are summarized as follows.
\begin{itemize}
 \item  We propose to model the dependencies of object attributes via a novel attribute knowledge graph. % a novel GCN based attribute learning model which can well capture and explore the dependencies of attributes for their prediction.
 \item  We propose a novel attribute knowledge graph convolutional network (GCN) based attribute learning model which can well capture and explore the dependencies of attributes for attribute representation and  prediction.
 \item We propose a strong discriminatory representation learning framework (AttKGCN) for the general object Re-ID tasks, which integrates both attribute learning and visual representation simultaneously and cooperatively in a unified network.
\end{itemize}
%Our HGC is illustrated by us with an example shown in Figure 1. We first extract the feature representation of parts in each layer and then construct a hierarchical graph for them. It can effectively enhance the discrimination ability of the final local features by learning the PH-GCN in this way and overcome the shortcomings of existing solutions (such as the lack of geometric structure informations between parts).

Extensive experiments on several benchmark datasets demonstrate % that the proposed AttKGCN approach outperforms the best state-of-the-art attribute-guided Re-ID methods.
 the effectiveness and benefits of the proposed AttKGCN approach.

%model figure
\begin{figure*}[ht]
\centering
\noindent\makebox[\textwidth][l] {
\includegraphics[width=1.0\textwidth]{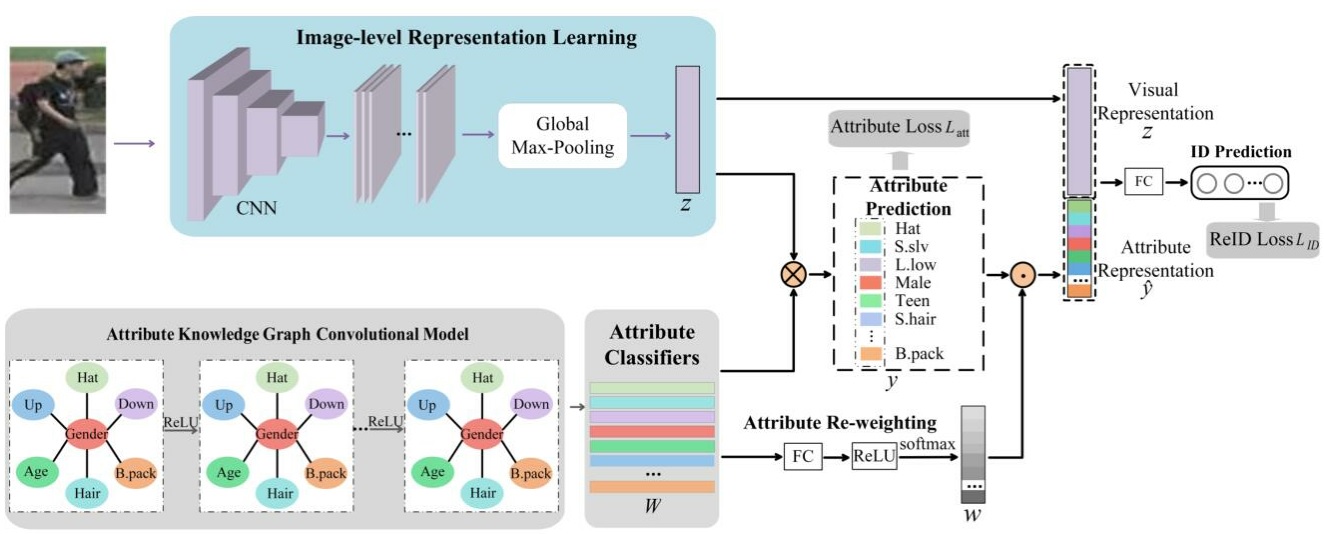}
}
\caption{Architecture of the proposed AttKGCN for person Re-ID, which contains three main parts, i.e., image-level representation learning, attribute knowledge graph convolutional module and attribute re-weighting. The backbone network can use any pre-trained general CNN models.} % Here, we use ResNet-50~\cite{he2016deep} as an example. }
\label{fig:model}
\end{figure*}

\section{Related Works}

\subsection{Attributed-based for Person Re-ID}

Recently, some studies have employed  attributes for person Re-ID~\cite{han2018attribute,lin2019improving, liu20183,li2019attributes, tay2019aanet} to improve Re-ID results.
Lin $et\ al.$~\cite{lin2019improving} annotate attributes for benchmark datasets Market-1501~\cite{Zheng_2015_ICCV} and DukeMTMC-reID~\cite{zheng2017unlabeled} manually, and propose an Attribute-Person Recognition (APR) network to conduct Re-ID embedding and pedestrian attributes prediction. %at the same time.
Han $et\ al.$~\cite{han2018attribute} propose Attribute-Aware Attention Model (A$^3$M) to jointly learn both local attribute and global identity feature representations together in an end-to-end manner.
Tay $et\ al.$~\cite{tay2019aanet} propose Attribute Attention Network (AANet) which integrates  person attributes and attribute attentions together for Re-ID problem.
%
% Tay $et\ al.$~\cite{tay2019aanet} propose Attribute Attention Network (AttNet) for person re-identification. It
%integrates  person attributes and attribute attention maps into a classification framework for Re-ID problem.
%
% to integrate body part information and the digital attribute information in an unified architecture.
%Layne $et\ al.$~\cite{Layne2012eccv} propose attribute-centric model to cooperate mid-level semantic attributes with low-level features for both identification and re-identification.
Li $et\ al.$~\cite{li2015multi} design a deep learning based single attribute recognition model (DeepSAR) to identify each attribute.
 They also present a deep learning (DeepMAR) for recognizing multiple attributes by exploiting the relationships among attributes.
Matsukawa $et\ al.$~\cite{matsukawa2016person} propose to define an attribute loss which is further combined with classification loss for Re-ID network training. %** for Ruse different attribute label for additional classification loss function, named attribute loss, which combined with general classification loss in order to enhance the discriminative ability of CNN.
%Zhang $et\ al.$~\cite{Zhang2018PersonRB} propose to conduct recognize attribute from the local regions of mid-level layers as auxiliary to improve the discrimination of high-level features.
Zhao $et\ al.$~\cite{zhao2019attribute} propose an attribute-driven method for  video person Re-ID problem. % by employing semantic attribute learning and recognition.

%\textbf{[add some more papers (two or more ) including all comparison (attribute re-id) works in experiments]}

Overall, the above attribute guided Re-ID approaches have demonstrated the benefits of integration of attribute learning and deeply visual representation for Re-ID problem.
However, they normally conduct attribute representation/learning individually which ignores the inherent co-occurrence information existing among different attributes.
To our best knowledge, this co-occurrence information has been less exploited or emphasized for Re-ID, although it has been mentioned (not utilized) in work~\cite{lin2019improving}.
Our aim in this paper is to further capture and exploit this co-occurrence information for attribute representation and Re-ID problem by employing the recently introduced Graph Convolutional Networks (GCNs)~\cite{kipf2016semi}.

%However, these work just use attributes-guide to lpearn a better feature representation.
%And it doesn't actually leverage attribute information in the datasets.

\subsection{Graph Convolutional Networks}

As an extension of CNNs from regular grid to irregular graph, graph convolutional networks (GCNs)~\cite{kipf2016semi,velickovic2017graph,Hu_2019,jiang2019semi,adaptive_GCN}
have been demonstrated very effectively in graph representation and learning.
Kipf $et\ al.$~\cite{kipf2016semi} propose to develop a simple Graph Convolutional Network (GCN) for graph semi-supervised learning.
Hamilton $et\ al.$~\cite{hamilton2017inductive} present a general inductive representation and learning framework for the representations of unseen nodes.
Veli{\v{c}}kovi{\'c} $et\ al.$~\cite{velickovic2017graph} propose Graph Attention Networks (GATs) for graph based semi-supervised learning.
The core GCNs is to conduct graph node representation and labeling by propagating messages on graph structure.
Recently, knowledge graph convolutional networks have been developed for zero-shot learning and multi-label recognition.
Wang $et\ al.$~\cite{wang2018zero} propose to employ GCN learning on a category knowledge graph to
predict the visual classifiers for unseen categories in zero-shot learning.
Kampffmeyer $et\ al.$~\cite{kampffmeyer2019rethinking} improve this work~\cite{wang2018zero} by further introducing a Dense Graph Propagation (DGP) module to exploit the hierarchical structure of knowledge graph.
Chen $et\ al.$~\cite{chen2019multi} propose multi-label GCN (ML-GCN) for image multi-label recognition.
 ML-GCN aims to learn inter-dependent object classifiers by employing GCN learning on an object label graph which encodes the correlation information among different labels.

%which leverage the probability of co-occurrence of labels to model the label correlation matrix and the word embeddings of multi-label as the input.
%
%designed a novel approach for zero-shot learning recognition by constructing a knowledge graph of relationship between the new category and familiar category and using the semantic embeddings of a category based on Graph Convolutional Network.
%
%
%And \cite{chen2019multi} proposed a GCN based model for multi-label image recognition named ML-GCN to learn inter-dependent object classifiers, which leverage the probability of co-occurrence of labels to model the label correlation matrix and the word embeddings of multi-label as the input.

Inspired by these works~\cite{wang2018zero,chen2019multi}, we propose to construct an attribute knowledge graph to model the co-occurrence information  among different person attributes and then employ a noval attribute knowledge graph convolutional network (AttKGCN) for person attribute prediction and Re-ID.

%Recently, GCNs have also been employed  to learn a set of visual classifiers.
%
%%
%GCNs aim to propagate messages on a graph structure. After message passing on the graph, the final
%node representation and labeling is determined based on its
%own appearance as well as its neighboring nodes information.
%
%
%Graph
%Since Kipf and Welling $et\ al.$~\cite{kipf2016semi} introduce Graph Convolutional Networks(GCNs) for semi-supervised learning on graph-structured data,
%which has attracted increasing attention not only in machine learning area~\cite{Hu_2019,adaptive_GCN,niepert2016learning,jiang2019semi}, but also in Computer vision~\cite{yan2019learning,gao2019graph,yan2018spatial,qi2018learning,guo2018neural}.
%For example, \cite{niepert2016learning} proposed a framework to learn representations for arbitrary graph through constructing locally connected neighborhoods from the input graphs.
%Jiang $et\ al.$~\cite{jiang2019semi} developed a novel Graph Learning-Convolutional Network(GLCN) to learn an optimal graph structure, which integrated graph learning and graph convolution into a unified framework.
%Hu $et\ al.$~\cite{Hu_2019} proposed a new deep Hierarchilcal Graph Convolutional Network(H-GCN), which repeatedly aggregated similar nodes in structure to hyper-nodes to form a coarsened graph and then refines it in order to restore the representation of each node for semi-supervised node classification.
%
%
%Kipf $et\ al.$~\cite{kipf2016semi} further introduce a more simple GCN Graph Convolutional Networks (GCNs) for semi-supervised learning on graph-structured data,

\section{Proposed Model}
% Overall introduction
In this section, we present our Attribute Knowledge Graph Convolutional Network (AttKGCN) for person attribute prediction and Re-ID tasks. %re-identification task.
The overall framework of AttKGCN is shown in Figure~\ref{fig:model}, which contains three main parts, i.e., 1) CNN based image feature extraction, 2) AttKGCN based attribute classifier generation and re-weighting and 3) integration of visual and attribute representations for final Re-ID task. %simulations  re-identification task.
In the following, we present the detail of these modules.
%AttGCN mainly contains two parts, one for attribute recognition task and the other for re-identification task.
%In image representation learning, Our purpose is to learn the features of an image.
%%Any CNN base models can be used in this sub-network.
%The GCN based classifier learning aim to learn inter-dependent attribute classifiers.

\subsection{Deeply visual representation}
\label{sec:3.1}
% simply introduction
Given an input person image $I$, we first extract its visual representation by using a deep feature extractor $\phi_{cnn}$.
The extractor $\phi_{cnn}$ can use any of pre-trained general CNN base models, such as ResNet-50~\cite{he2016deep}, VGG~\cite{simonyan2014very}, or some  deeply person-specific feature extraction models, such as PCB~\cite{sun2018beyond}, HPM~\cite{fu2019horizontal}, etc.
For convenience, here we use ResNet-50~\cite{he2016deep} as an example for description.
In this case, we can first obtain 2048$\times$24$\times$8 feature maps from the $"conv5{\_}x"$ layer when we resize an input image $I$ as 384$\times$128.
Then, we obtain the image-level feature $z\in \mathbb{R}^d$ ($d=2048$) by further employing global max$\_$pooling operation as follow,
\begin{equation}
z = f_{\mathrm{GMP}}\left (\phi_{cnn}\left ( I;\Theta _{cnn}\right )\right )%\in R^{D}
\label{equ:1}
\end{equation}
where $f_{\mathrm{GMP}}\left( \cdot \right)$ indicates the global max-pooling operation, $\Theta _{cnn}$ denotes the parameters of CNN model. % and $d$ denotes feature dimension of output, which is 2048 in our experiments.

\subsection{AttKGCN based attribute learning}
\label{sec:3.2}
The aim of the proposed AttKGCN is to learn inter-dependent attribute classifiers via GCN based mapping function.
In order to do so, we first construct an attribute knowledge graph (AttKG).
Then, we propose to develop an attribute knowledge graph convolutional learning module for attribute representation and prediction.
Finally, we propose to integrate attribute learning and visual representation together in a unified manner for the final Re-ID task (see in section~\ref{sec:3.3}).
% The overall network is shown in Figure 2.

%Our model builds upon the Graph Convolutional Network by leveraging the prior information of attributes in the datasets.
%In this secton, we first show the construction of our attribute graph and the propagation details of our model on GCN.
%Then we present our proposed AttGCN how to make attribute predictions.
%Finally, how the proposed AttGCN we designed applies to Re-ID.

%\subsubsection{Attribute knowledge graph construction}

\begin{figure*}[ht]
\centering
\noindent\makebox[\textwidth][l] {
\includegraphics[height=8cm,width=17.5cm]{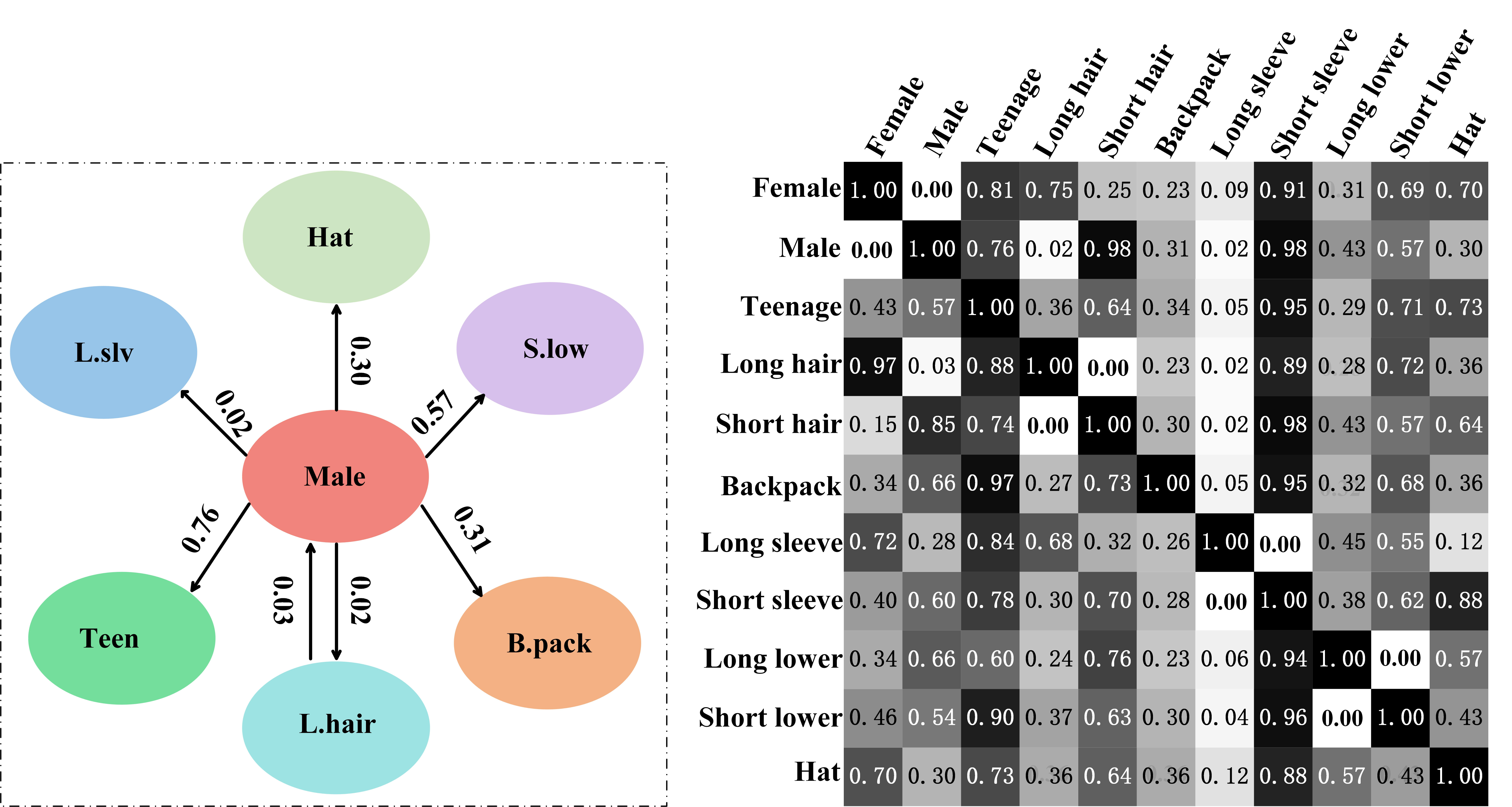}
}
\caption{An example of the proposed AttKG obtained from Market-1501~\cite{Zheng_2015_ICCV} dataset. In this figure, we use the depth of the color to indicate the strength of the correlation between the two attributes. $'L.hair'$, $'B.pack'$, $'L.slv'$, $'S.low'$, $'Teen'$ denotes $'Long\ hair'$, $'Backpack'$, $'Long\ sleeve'$, $'Short\ lower'$, $'Teenage'$, respectively.}
\label{fig:kg}
\end{figure*}

\textbf{Attribute Knowledge Graph Construction.}
%The core of GCN based learning is to propagate information on a given graph which encodes the
%correlation among different data points.
In order to employ GCN learning for attribute representation and prediction, we need to first
construct an attribute graph to represent the correlation among different attributes.
Specifically, as shown in Figure~\ref{fig:kg}, we construct an attribute knowledge graph (AttKG) $G(A,P)$ as follows.
Each node $v_i, i=1,2\cdots c$ in AttKG denotes a specific attribute $A_i \in A$, such as \emph{Female}, \emph{Hat}, etc., which can be  described via a semantic embedding vector such as word embedding. Here $c$ denotes the number of different attributes.
An edge $e_{ij}$ exists between each node pair in AttKG which encodes the co-occurrence relationship $P_{ij}$ between attribute $A_i$ and $A_j$.
Similar to work~\cite{chen2019multi}, we define this co-occurrence relationship $P_{ij}$ via conditional probability $p\left ( A_j \mid A_i\right )$ which denotes the probability of occurrence of attribute $A_j$ when attribute $A_i$ appears. %, which is defined as $P\left ( A_j \mid A_i\right )$.
Given a training dataset, the conditional probability $P_{ij}=p\left ( A_j \mid A_i\right)$ is estimated as,
\begin{equation}
P_{ij} = p\left ( A_j \mid A_i\right ) = \frac{m_{ij}}{n_i}  \quad\quad   i, j = 1, 2 \cdots c
\end{equation}
where $m_{ij}$ denotes the co-occurring number of attribute $A_i$ and $A_j$, $i.e.$, the number of person image in training dataset that contains both attribute $A_i$ and $A_j$, and $n_i$ denotes the number of attribute $A_i$, $i.e.$, the number of person image that contains attribute $A_i$ in the training dataset.
Obviously, the co-occurrence probability matrix $P \in \mathbb{R}^{c\times c}$ is unsymmetrical, which indicates the unsymmetry of our attribute knowledge graph $G(A, P)$.
Figure~\ref{fig:kg} shows an example of the proposed AttKG which is constructed based on the attributes of Market-1501~\cite{Zheng_2015_ICCV} dataset.

%Adj figure

\textbf{AttKGCN for Attribute Representation.}
Given an input AttKG $G(A,P)$, we adopt a $L$-layer GCN module for attribute representation and learning.
Formally, given ${P}\in \mathbb{R}^{c\times c}$ and the $l$-layer representation ${H}^{(l)}\in \mathbb{R}^{c\times d_l}$, we propose to conduct the layer-wise propagation  as
%The main purpose of this module is to extract structural information for image $Ip$ of pedestrian retrieval task.
%We employ the PGCN module to extract high-level feature representation for nodes of the intra-graphs $G(F, A)$.
%In other words, we process multiple-layer PGCN propagation that is formulated as
%\begin{center}
%
%Formally, given a feature descriptions $H^{(l)} \in R^{C \times d_0}$ as the other, note that $C$ represents the number of nodes and $d_0$ denotes the dimensionality of node features.
%After applying \cite{kipf2016semi} of the same convolution operations, the node features is updated as $H^{(l+1)} \in R^{C \times {d}'}$, the propagation of each GCN layer can be represented as follow

\begin{equation}
H^{(l+1)} = \sigma(\tilde{P} H^{(l)} \Theta^{(l)})
\end{equation}
where $l = 0, 1, ..., L-1$, and $\tilde{P}$ indicates the normalized matrix $P$.
$\sigma(\cdot)$ is an activation function, such as ReLU function $\mathrm{ReLU}(\cdot) = \max(0,\cdot)$. % in our experiment.
$\Theta^{(l)}$ denotes the layer-wise  trainable weight matrix.

The first input feature representation $H^{(0)}$ of AttKGCN can be set as the semantic embeddings of attributes. % can be taken from any prior attribute representation,
%such as, word embeddings.
In this paper, we just simply set $H^{(0)}=I$ which can also obtain desired learning results, where $I$ denotes an identity matrix with proper size.
The main purpose of AttKGCN module is to learn a set of inter-dependent classifiers for attributes which are combined with the pre-trained deeply visual  representation for person attribute prediction.
Thus, the final output of AttKGCN is a regression matrix  $U  = H^{(L)} \in \mathbb{R}^{c\times d}$ where
$d$ denotes the dimensionality of the image feature (Eq.~(\ref{equ:1})) and $c$ denotes the number of different attributes. %classifier to the $i$-th attribute $A_i$.
Overall, we can summarize our AttKGCN module as
\begin{equation}
U =  H^{(L)} = \phi_{gcn} (P,\Theta_{gcn})      % {y} =({y}_1, {y}_2\cdots {y}_c) = \mathrm{softmax}(Wz)
\label{equ:4}
\end{equation}
where $\Theta_{gcn} = \{\Theta^{(0)}, \Theta^{(1)}\cdots \Theta^{(L-1)}\}$ denotes the collection of network weight matrices.
In the following, we will apply the learned regressor $U$ for attribute prediction and re-weighting tasks.

% \subsubsection{Attribute Prediction and Re-weighting}
\textbf{Attribute Prediction and Re-weighting.}
The final output of the above AttKGCN is a regression matrix  $U\in \mathbb{R}^{c\times d}=\{U_i\}^c_{i=1}$ where each row $U_i\in \mathbb{R}^{1\times d}$ denotes the classifier to the $i$-th attribute $A_i$.
Specifically, given an input person visual representation $z\in \mathbb{R}^{d\times 1}$, we can obtain the predicted scores of all attributes for this person by applying the learned regressor $U$ as follows,
\begin{equation}
{y} =({y}_1, {y}_2\cdots {y}_c)^T = \mathrm{softmax}(Uz), \ \ \ y_i = U_iz
\end{equation}
where ${y}_i$ denotes the predicted score of the $i$-th attribute, and
$\mathrm{softmax} (\cdot)$ is the softmax function which guarantees the output attribute scores satisfying the probability
condition, $i.e.$, $\sum^c_{i=1}{y}_i=1, {y}_i\geq 0$.
Here, we adopt cross-entropy loss function~\cite{de2005tutorial} for attribute prediction,
%Finally, the attribute loss can be computed as follow
\begin{equation}
\mathcal{L}_{att}=-\sum_{j=1}^{c} y^{gt}_i \log({y}_i)% \left ( j\right ) \log \left ( p\left ( j\right )\right )
\end{equation}
where $y^{gt}=(y^{gt}_1,y^{gt}_2\cdots y^{gt}_c)^T$ denotes the ground-truth  attribute vector of input image $I$, $i.e.$, $y^{gt}_i =1$ indicates that  image $I$ has attribute $A_i$, and $y^{gt}_i =0$ otherwise. %, i.e, $q\left ( y_C\right ) = 1$ and $q\left ( j\right ) = 0$ for all $j \neq y_C$.

Moreover, in order to utilize the attribute information for Re-ID  more compactly,
it is also desirable to recalibrate the strengths of different attributes for person
attribute representation. %image Re-ID task.
This motivates us to develop an attribute re-weighting (AttRW) module.
In this paper, we formulate AttRW task as  node labeling/weighting on attribute graph $G(A,P)$ and implement it
using GCN learning.
Given attribute knowledge graph $G(A,P)$,  AttRW aims to obtain
weights $w=(w_1, w_2, \cdots w_c)\in \mathbb{R}^{c}$ for different attributes by using
\begin{align}
w = \mathrm{Sigmoid} \big( f_{\mathrm{FC}}(\phi_{\mathrm{gcn}}(P,\tilde{\Theta}_{gcn}),\theta)\big)
\end{align}
where $f_{\mathrm{FC}}$ denotes a single layer neural network with parameter $\theta$.
The purpose of sigmoid function is used to guarantee the nonnegativity of $w$.

\noindent \textbf{Remark.} Here, one can either utilize the same GCN module used in Eq.~(\ref{equ:4}) ($i.e.$, setting $\Theta_{gcn} = \tilde{\Theta}_{gcn}$) or design a different new GCN module. In our experiments, we use the former setting for complexity consideration. %the same , i.e., $\Theta_{gcn} = \Theta'_{gcn}$

By applying attribute weights $w$,
we can obtain a kind of weighted attribute representation $\hat{y}$ for Re-ID task as
%Hence, the original attribute prediction $\hat{y}$ is converted into a new prediction score by the the attribute re-weight module as follow
\begin{equation}
\hat{y} = w \odot {y}
\end{equation}
where $\odot$ denotes the element-wise multiplication operation between vector $w$ and $y$.

\subsection{Re-identification}
\label{sec:3.3}
% ½éÉÜÈçºÎ½«attribute prediction incorproaed with image cnn feature for reid

We propose to integrate both visual and re-weighted attribute representations for person Re-ID.
%
%For the final re-identification, we first concatenate
%In order to further identity classification,
To do so, we first develop a visual-attribute representation for the input person image $I$ by concatenating its deeply learned image representation $z$ with its corresponding re-weighted attribute prediction vector $\hat{y}$. Then, we adopt a FC layer followed by softmax operation to predict the ID label of the person, $i.e.$,
\begin{align}
p = \mathrm{softmax} \big( f_{\mathrm{FC}}( [z\parallel\hat{y}],\theta)\big)
\end{align}
where $\|$ denotes the concatenation operation. Here, we adopt cross-entropy loss~\cite{de2005tutorial} function $\mathcal{L}_{id}$ for identity prediction.
Therefore, the final overall loss function is
% Thus, we define the overall objective function as follow
\begin{equation}
\mathcal {L}  =  \mathcal{L}_{att} + \lambda \mathcal {L}_{id}
\label{equ:10}
\end{equation}
where $\lambda$ is a hyper-parameter to balance  identification loss and attribute prediction loss. % $L_{ID}$ denotes the identity classification loss~\cite{de2005tutorial}, which use cross entropy loss to optimize, and the details can be referred to APR~\cite{lin2019improving}.

\begin{table*}[]
\caption{Comparison with state of the art on the Market-1501~\cite{Zheng_2015_ICCV} and DukeMTMC-reID~\cite{zheng2017unlabeled}. \textbf{C:} the methods of CNN-based on re-id. \textbf{A:} the methods of Attribute-based on re-id. The {\color{red}\textbf{red}}/{\color{blue}\textbf{blue}} denotes the 1$^{st}$/2$^{nd}$ best results, respectively.}
\begin{center}
\vspace{-0.35cm}
\begin{tabular}{l|l|l|l|l|l|l|c|c|c|c}
\hline
\multicolumn{2}{c|}{\multirow{2}{*}{Method}}            & \multirow{2}{*}{Reference} & \multicolumn{4}{c|}{Market1501}   & \multicolumn{4}{c}{DukeMTMC-reID} \\ \cline{4-11}
\multicolumn{2}{c|}{}                                   &                            & Rank-1 & Rank-5 & Rank-10 & mAP   & Rank-1  & Rank-5 & Rank-10 & mAP   \\ \hline\hline
\multirow{14}{*}{C}                      & FD-GAN~\cite{ge2018fd}            & NIPS2018                   & 90.5   & -      & -       & 77.7  & 80.0    & -      & -       & 64.5  \\ \cline{2-11}
                                         %& CamStyle~\cite{zhong2018camstyle} & CVPR2018                   & 88.1  & 95.1   & 97.0    & 68.7 & 75.2   & 84.6   & 87.9    & 53.4 \\ \cline{2-11}
                                         %& PSE~\cite{saquib2018pose}         & CVPR2018                   & 87.7   & -      & -       & 69.0  & 79.8    & -      & -       & 62.0  \\ \cline{2-11}
                                         & SafeNet~\cite{yuan2018safenet}    & IJCAI2018                  & 90.2   & -      & -       & 72.7  & 82.7    & -      & -       & 57.0  \\ \cline{2-11}
                                         & MGCAM~\cite{song2018mask}         & CVPR2018                   & 83.7  & -      & -       & 74.33 & -       & -      & -       & -     \\ \cline{2-11}
                                         & PCB~\cite{sun2018beyond}          & ECCV2018                   & 92.4   & 97.0   & 97.9    & 77.3  & 81.9    & 89.4   & 91.6    & 65.3  \\ \cline{2-11}
                                         & SGGNN~\cite{shen2018person}       & ECCV2018                   & 92.3   & 96.1   & 97.4    & 82.8  & 81.1    & 88.4   & 91.2    & 68.2  \\ \cline{2-11}
                                         & MGAT~\cite{bao2019masked}         & CVPRW2019                  & 91.5   & 97.2   & 98.0    & 76.5  & -       & -      & -       & -     \\ \cline{2-11}
                                         & VPM~\cite{sun2019perceive}        & CVPR2019                   & 93.0   & {\color{blue}\textbf{97.8}}   & {\color{red}\textbf{98.8}}    & 80.8  & 83.6    & 91.7   & 94.2    & 72.6  \\ \cline{2-11}
                                         & CASN~\cite{zheng2019re}           & CVPR2019                   & {\color{blue}\textbf{94.4}}   & -      & -       & 82.8  & {\color{blue}\textbf{87.7}}    & -      & -       & 73.7  \\ \cline{2-11}
                                         & HPM~\cite{fu2019horizontal}       & AAAI2019                   & 94.2   & 97.5   & 98.5    & 82.7  & 86.6    & {\color{blue}\textbf{93.0}}   & {\color{blue}\textbf{95.1}}    & {\color{blue}\textbf{74.3}}  \\ \cline{2-11}
                                         & TBN+~\cite{li2019pedestrian}      & ICME2019                   & 93.2   & -      & -       & 83.0  & 85.5    & -      & -       & 73.0  \\ \cline{2-11}
                                         & \cite{Zhou_2019}                  & ICCV2019                   & 96.1   & -      & -       & {\color{blue}\textbf{84.7}}  & 86.3    & -      & -       & 73.1  \\ \hline
\multicolumn{1}{c|}{\multirow{6}{*}{A}} & ACRN~\cite{schumann2017person}     & CVPR2017                   & 83.6  & 92.6  & 95.3   & 62.6 & 72.5   & 84.7  & 88.8   & 51.9 \\ \cline{2-11}
\multicolumn{1}{c|}{}                   & APR~\cite{lin2019improving}        & PR2019                     & 87.0  & 95.1  & 96.4   & 66.8 & 73.9   & -      & -       & 55.5 \\ \cline{2-11}
\multicolumn{1}{c|}{}                   & APDR~\cite{li2019attributes}       & PR2019                     & 93.1   & 97.2   & 98.2    & 80.1  & 84.3    & 92.4   & 94.7    & 69.7  \\ \cline{2-11}
\multicolumn{1}{c|}{}                   & MLFN~\cite{chang2018multi}         & CVPR2018                   & 87.0  & 95.1  & 96.4   & 66.8 & 73.9   & -      & -       & 55.5 \\ \cline{2-11}
\multicolumn{1}{c|}{}                   & A$^{3}$M~\cite{han2018attribute}   & MM2018                     & 86.5  & 95.1  & 97.0   & 68.9 & -       & -      & -       & -     \\ \cline{2-11}
%\multicolumn{1}{c|}{}                   & \cite{Zhang2018PersonRB}           & ACML2018         & 87.8  & 94.8  & 97.0   & 70.0 & 79.6   & -      & -       & 61.4   \\ \cline{2-11}
\multicolumn{1}{c|}{}                   & AANet~\cite{tay2019aanet}          & CVPR2019                   & 93.8  & -      & 98.5   & 82.4 & 86.4   & -      & -       & 72.5 \\ \hline\hline
%\multicolumn{2}{c|}{\textbf{Baseline1}}& -                            & 67.9  & 85.1  & 89.3   & 43.8 & 55.5 & 71.5 & 78.1 &36.0   \\ \hline
%\multicolumn{2}{c|}{\textbf{AttKGCN}}                            & -                          & 73.0 & 88.2 & 92.2 &50.0       &63.9        & 76.8         &81.6   & 41.1     \\ \hline
%\multicolumn{2}{c|}{\textbf{Baseline2}}& -                          & 93.3   & 97.6   & 98.5    & 81.7  & 87.4    & {\color{blue}93.3}   & 95.0      & {\color{blue}74.4} \\ \hline
\multicolumn{2}{c|}{\textbf{AttKGCN}}                               & -                          & {\color{red}\textbf{94.4}}   & {\color{red}\textbf{98.0}}   & {\color{blue}\textbf{98.7}}    & {\color{red}\textbf{85.5}}  & {\color{red}\textbf{87.8}}    & {\color{red}\textbf{94.4}}   & {\color{red}\textbf{95.7}}      & {\color{red}\textbf{77.4}}  \\ \hline
\end{tabular}
\end{center}
\label{table:1}
\end{table*}

% Market1501 attribute predipction
\begin{table*}[]
\begin{center}
\caption{Results of our approach for attribute recognition on Market-1501~\cite{Zheng_2015_ICCV}. 'Avg' is our average accuracy to indicate overall attribute prediction accuracy. 'B.pack', 'H.bag', 'C.down', 'C.up', 'S.clth', 'L.low', 'L.slv' denotes $'Backpack'$, $'Handbag'$, $'the\ color\ of\ lower$-$body\ clothing'$, $'the\ color\ of\ upper$-$body\ clothing'$, $'the\ type\ of\ lower$-$body\ clothing'$, $'the\ length\ of\ lower$-$body\ clothing'$, $'sleeve\ length'$, respectively.}
\begin{tabular}{l|c|c|c|c|c|c|c|c|c|c|c|c|c}
\hline
Method & Age  & B.pack & Bag  & H.bag & C.down & C.up & S.clth & L.low & L.slv & Hair & Hat  & Gender & Avg  \\  \hline
APR~\cite{lin2019improving} & 88.6 & 84.9   & 76.4 & 90.4  & 73.8   & 74.0 & 92.8   & 93.7  & 93.6  & 84.4 & 97.1 & 88.9   & 86.6 \\  \hline
AANet~\cite{tay2019aanet}   & 88.2 & 87.7   & 79.7 & 89.6  & 70.81   & 77.1 & 94.8   & 94.2  & 94.4  & 86.5 & 98.0 & 92.3   & 87.8 \\  \hline
AttKGCN & 88.9 & 90.0   & 89.6 & 89.3  & 90.1   & 88.5 & 94.0  & 89.8  & 89.0  & 90.1 & 89.5 & 89.4   & 89.8 \\  \hline
\end{tabular}
\label{table:2}
\end{center}
\end{table*}

\section{Experiments}
%In this section, we evaluate our proposed AttKGCN model on two popular person Re-ID benchmarks, including Market-1501~\cite{Zheng_2015_ICCV} and DukeMTMC-reID~\cite{zheng2017unlabeled}.
%First, we briefly describe the datasets and the implementation details of our AttKGCN framework.
%Second, we analyze our method quantitatively and qualitatively.
%After that, some parameter analysis results and ablation study will be presented.
%More detail can see below.

%We will first introduce the datasets and evaluation metric, and then briefly describe the implementation dpetails of our AttKGCN framework.
%In comparisons with state-of-the-art methods, We divide it into two tasks: person Re-id and attribute recognition.
%And there are many Re-ID methods.
%We mainly elaborate on our baseline, CNN-based and attribute-based methods, respectively.
%After that, some parameter analysis results and ablation study will be presented.

\subsection{Datasets and Settings}

\noindent \textbf{Market-1501}~\cite{Zheng_2015_ICCV} dataset contains 32668 images of 1501 persons which are observed under six camera viewpoints.
We follow the standard training and evaluation protocol where 750 identities are selected for testing and 751 identities for training.
Deformable Part Model (DPM)~\cite{forsyth2014object} is used as the person detector.
For each identity in this dataset, 27 attribute labels are annotated by~\cite{lin2019improving}.
%And ~\cite{lin2019improving} annotates 27 attributes in the identity level for this dataset.
%We use all attributes in this work.

\noindent \textbf{DukeMTMC-reID}~\cite{zheng2017unlabeled} dataset is a subset of DukeMTMC dataset~\cite{ristani2016performance} which contains 1812 identities observed from 8 different camera viewpoints.
It is divided into 16522 training images of 702 identities and 19889 test images of 702 identities.
The evaluation protocol is same as that of on Market-1501 dataset~\cite{Zheng_2015_ICCV}.
For each identity in this dataset, 23 attribute labels are annotated by~\cite{lin2019improving}.

%~\cite{lin2019improving} also annotated this dataset, but with 23 attributes in the identity level.

 {\textbf{Evaluation Metrics.}}
Following many previous works~\cite{sun2018beyond,zhong2017re}, we use both Cumulative Matching Characteristic (CMC) (as rank-1, rank-5 and rank-10) and mean Average Precision (mAP) for evaluation, where mAP denotes the mean value of average precision across all queries.
For attribute recognition, we calculate the classification accuracy for each attribute as well as the overall averaged accuracy of all these attribute predictions.
%For the attribute task, we use the gallery images as the testing set to compute the classification for each attribute.
%But for
When testing attribute prediction on Market-1501~\cite{Zheng_2015_ICCV}, we omit the distractor (background) and junks images because they do not have attribute labels~\cite{lin2019improving}.
%
% since  didn't annotate the distractor (background) and junks images,
%
% elide them when we test the attribute prediction.
%We report our overall attribute prediction accuracy via computing the averaged accuracy of all these attribute predictions as ~\cite{lin2019improving}.
%

\subsection{Implementation details}
\label{sec:4.2}
As discussed in section~\ref{sec:3.1},
for each person image, we extract its visual representation by using
any person specific deep feature extraction models.
In our experiments, we implement Horizontal Pyramid Matching (HPM) network~\cite{fu2019horizontal} (Baseline model)
 and use it as our feature extractor $\phi_{cnn}$.
 %
% During training, we augment the training images with horizontal flipping, nomalization and randomly erase.
%
In our experiment, the batch size is set to 90 and the epoch number is set to 120.
The  learning rate is set to 0.1 and 0.001 for the feature extraction module and our attribute GCN module, respectively. % and set to 0.001 for GCN model.
%The learning rate for HPM is not fixed. For the first 60 epochs, the learning rate decayed to 0.01 after 40 epochs and changed to 0.001 in the last 60 epochs.
%
We train the whole network by using stochastic gradient descent (SGD)~\cite{bottou2010large} in each mini-batch.
The whole training process takes about two hours.

\subsection{Experiment results}
\label{sec:4.3}
%\subsubsection{Evaluation on person Re-ID}
%
% To verify the effectiveness of the proposed AttKGCN method in re-ID task, we conduct experiments on two large-scale person re-ID datasets Market-1501~\cite{Zheng_2015_ICCV} and DukeMTMC-reID~\cite{zheng2017unlabeled}.

We compare our AttKGCN with some recent related state-of-art methods which mainly includes both recent CNN-based Re-ID method FD-GAN~\cite{ge2018fd}, TBN+~\cite{li2019pedestrian}, VPM~\cite{sun2019perceive}, SGGNN~\cite{shen2018person}, PCB~\cite{sun2018beyond}, HPM~\cite{fu2019horizontal} and attribute-based method APDR\cite{li2019attributes}, APR~\cite{lin2019improving}, AANet~\cite{tay2019aanet} and MLFN~\cite{chang2018multi}.
%
%We mainly compare our AttKGCN with some recent related state-of-art methods that CNN-based and attribute-based, respectively.
%For example, CNN-based methods: FD-GAN~\cite{ge2018fd}, TBN+~\cite{li2019pedestrian}, VPM~\cite{sun2019perceive}, SGGNN~\cite{shen2018person}, PCB~\cite{sun2018beyond} and so on.
%Attribute-base methods: APDR\cite{li2019attributes}, APR~\cite{lin2019improving}, AANet~\cite{tay2019aanet} and MLFN~\cite{chang2018multi} and so on.
%
Table~\ref{table:1} summarizes the comparison results on Market-1501~\cite{Zheng_2015_ICCV} and DukeMTMC-reID~\cite{zheng2017unlabeled} datasets.
% For fair comparison, no post-processing such as re-ranking~\cite{zhong2017re} is used for our AttKGCN methods.
Overall, AttKGCN generally obtains the best results on two benchmarks.
More concretely, we can note the following aspects.
%Finally, we implement our method with two versions, i.e., PH-GCN and PH-GCN+Re-ranking.
%PH-GCN+Re-ranking further uses re-ranking~\cite{zhong2017re} approach to improve the Re-ID results.

\emph{Comparison with CNN-based Methods}.
We select some representative CNN-based Re-ID methods which mainly include traditional CNN methods, augmented dataset methods, part-based methods and graph-based methods, etc.
% AttKGCN gains rank-1 = 94.4$\%$ , mAP = 85.5$\%$
For example, on Market-1501, comparing with part-based convolutional baseline (PCB)~\cite{sun2018beyond}, AttKGCN has 8.2$\%$ and 2.0$\%$ improvements on mAP and rank-1, respectively.
Comparing with Horizontal Pyramid Matching (HPM), AttKGCN result improves 2.8$\%$ and 0.2$\%$ on mAP and rank-1, respectively.
On DukeMTMC-reID~\cite{zheng2017unlabeled}, % AttKGCN gains rank-1 = 77.4$\%$ , mAP = 87.8$\%$.
comparing with PCB~\cite{sun2018beyond} and HPM~\cite{fu2019horizontal},
AttKGCN has 5.9$\%$ and 1.2$\%$ improvements on rank-1 and 12.1$\%$ and 3.1$\%$ improvements on mAP, respectively.
These clearly demonstrate the effectiveness of AttKGCN to enhance the discriminatory ability of part-based Re-ID model by incorporating our attribute knowledge graph learning architecture.
% and thus Re-ID results.
%The rank-1 rises by 5.9$\%$ and 1.2$\%$ correspondingly, and the mAP rises by 12.1$\%$ and 3.1$\%$ correspondingly by comparing with PCB~\cite{sun2018beyond} and HPM~\cite{fu2019horizontal}.
%It thoroughly demonstrates the effectiveness of our AttKGCN model on part-based learning framework.

%rank
\begin{figure}[ht]
\centering
\noindent\makebox[\textwidth][l] {
\includegraphics[height=6cm,width=8.2cm]{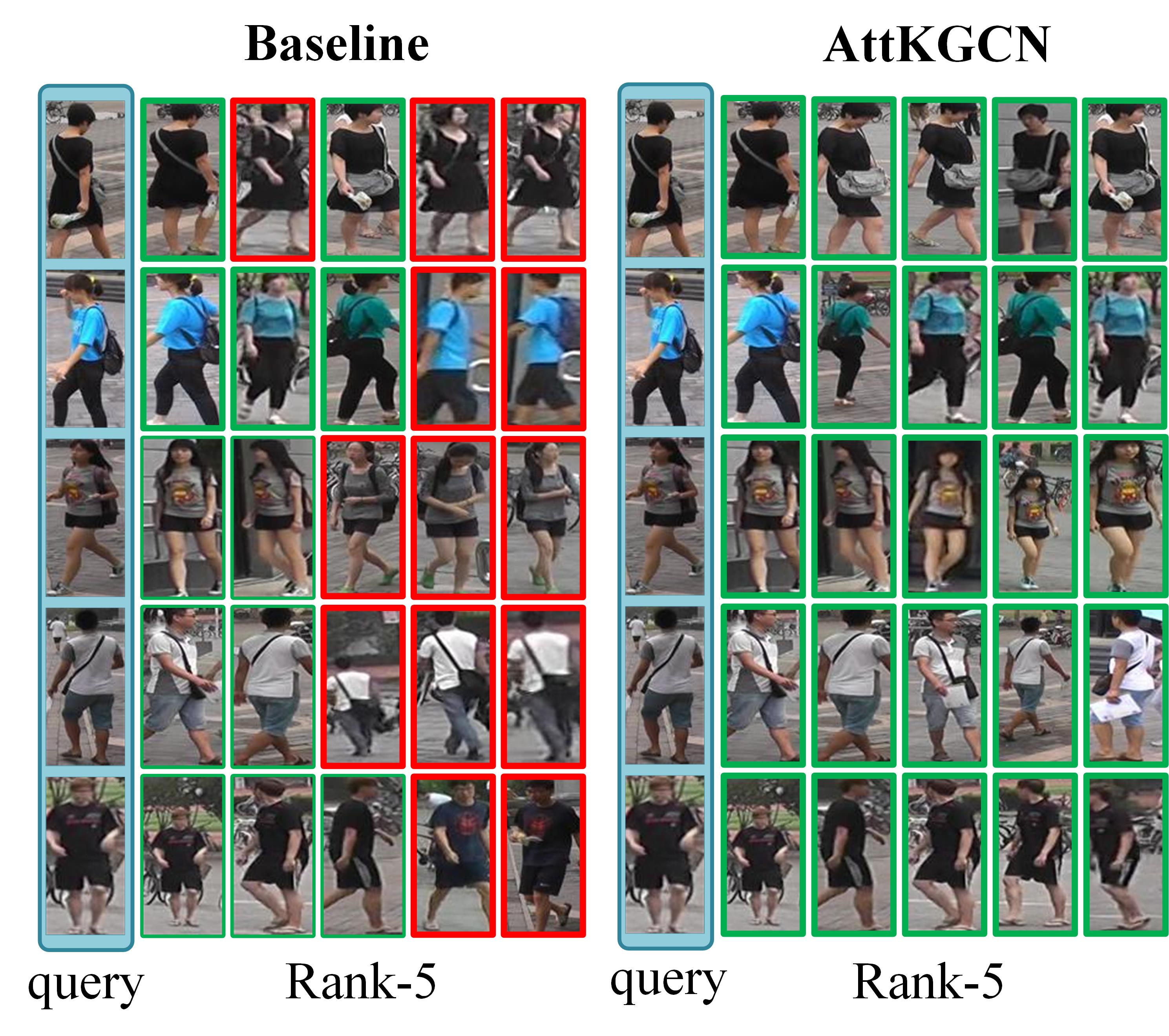}
}
\caption{Rank-5 results for some queries returned by Baseline and AttKGCN on Market-1501~\cite{Zheng_2015_ICCV}, respectively. Green boundary indicates true positive samples, and red represents false positive samples.}
\label{fig:rank}
\end{figure}

%t-SNE
\begin{figure}[ht]
\centering
\noindent\makebox[\textwidth][l] {
\includegraphics[height=4cm,width=8.2cm]{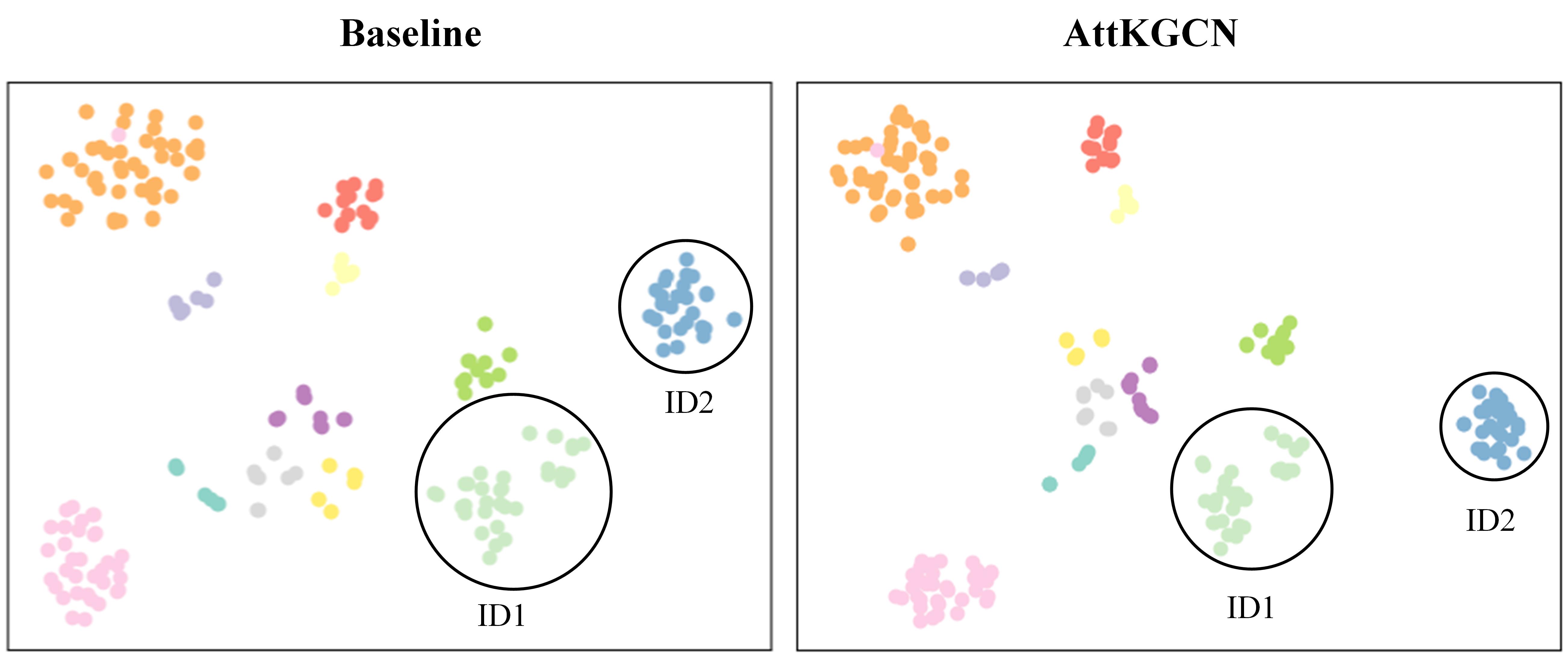}
}
\caption{2D t-SNE~\cite{van2008visualizing} visualization of  feature representations learned by Baseline and  AttKGCN on Market-1501~\cite{Zheng_2015_ICCV}, respectively.}
\label{fig:tsne}
\end{figure}

\emph{Comparison with Attribute-based Methods}.
We compare AttKGCN with recent attribute-based methods including ACRN~\cite{schumann2017person}, A${^3}$M~\cite{han2018attribute}, APR~\cite{lin2019improving} and AANet~\cite{tay2019aanet}.
Table~\ref{table:1} summarizes the comparison results.
Here, we can note that, comparing with recent state-of-the-art attribute-based Re-ID method AANet~\cite{tay2019aanet}, AttKGCN gains 3.1$\%$ and 0.6$\%$ improvements on mAP and rank-1, respectively on Market-1501~\cite{Zheng_2015_ICCV} dataset. %, respectively.
On DukeMTMC-reID~\cite{zheng2017unlabeled}, we also gain 4.9$\%$ and 1.4$\%$ improvements over AANet~\cite{tay2019aanet} on mAP and rank-1 measurements.
% For DukeMTMC-reID~\cite{zheng2017unlabeled}, mAP improved 4.9$\%$ and rank-1 increased 1.4$\%$ .

%
% on two benchmarks.
%As far as we know, we realize the best performance of attribute based method.
%We present some representative attribute-based person re-identification methods,

\emph{Qualitative Visualization}.
Figure~\ref{fig:rank} shows the top-5 ranking list of Re-ID result for some query images on Market-1501~\cite{Zheng_2015_ICCV}. %, which experiments on Baseline and our proposed AttKGCN, respectively.
Intuitively, one can note that AttKGCN can find more true positives than Baseline model\footnote{ It has the same network setting with AttKGCN but without attribute representation/learning module.}.

\emph{Representation Visualization}.
Figure~\ref{fig:tsne} demonstrates 2D  visualizations of the learned final output representation of Baseline network and AttKGCN on  Market-1501 dataset, respectively. Here, we only show 11 identities and different colors denote different identities. Intuitively, one can observe
that the persons'  representations of AttKGCN are distributed more clearly than Baseline, which further demonstrates
the more discriminatory of AttKGCN by incorporating attribute representation in person representation. % of the proposed RobustGCN on graph data representation.

% The above results indicate the effectiveness of our proposed AttKGCN by quantitatively and qualitatively.
%Here, we visualize the learned features for Re-ID in order to show the effect of our method on classification.
%We randomly selected eleven identities on the test set of Market-1501~\cite{Zheng_2015_ICCV} dataset to experiment, as shown in Figure~\ref{fig:tsne}.
%One color represents one identity.
%%It is clearly to see that the classifiers learned by our proposed AttKGCN minish the difference of intra-class.
%It is clearly to see that our AttKGCN minishs the intra-class distances of cycle ID1 and ID2 and the others also lie in a compact region.

\subsection{Evaluation on Attribute Recognition}
To evaluate the effectiveness of the proposed GCN based attribute prediction, we test attribute recognition/prediction on  Market-1501~\cite{Zheng_2015_ICCV} dataset. %and DukeMTMC-reID~\cite{zheng2017unlabeled}.
%For Market-1501 dataset, which is annotated 27 attributes by Lin~$et\ al.$~\cite{lin2019improving}.
%And with 23 attribute on DukeMTMC-reID.
Table~\ref{table:2} summarizes the comparison results.
Overall, AttKGCN performs better than the competing method APR~\cite{lin2019improving} and AANet~\cite{tay2019aanet} on most attributes and obtains the highest prediction accuracy on average results.
This clearly demonstrates the
effectiveness of the proposed AttKGCN on person attribute recognition.

%% The results are show in Table 2 and Table 3.
%By comparing with the results of APR~\cite{lin2019improving}, our main summary is as follows:
%First, on two datasets, our proposed AttKGCN network improve the overall attribute recognition accuracy to a certain extent.
%The improvements are 3.8$\%$ and 0.03$\%$ on Market-1501 and DukeMTMC-reID, respectively.
%So to sum up, we introduce some degree of complementary information such as attribute for object classification can help training a more discriminative ability model.
%Also, as far as we know, up to now, we achieve the best attribute recognition result on Market-1501 and DukeMTMC-reID.
%Second, we observe that the recognition rate of each attribute is small difference.
%Overall, the prediction accuracy of each attribute is mostly improving on Market-1501.
%But there are few attribute recognition rates have decreased, such as $"Handbag"$, $"L.low"$, $"L.slv"$ and $"Hat"$.
%So it fully demonstrates the effectiveness of our proposed AttKGCN in attribute recognition task and the stability of prediction for each attribute.

\subsection{Parameter Analysis}
\label{sec:4.4}
\begin{figure}
\centering
\subfigure[mAP]{
\label{figa} %% label for first subfigure
\includegraphics[width=1.59in]{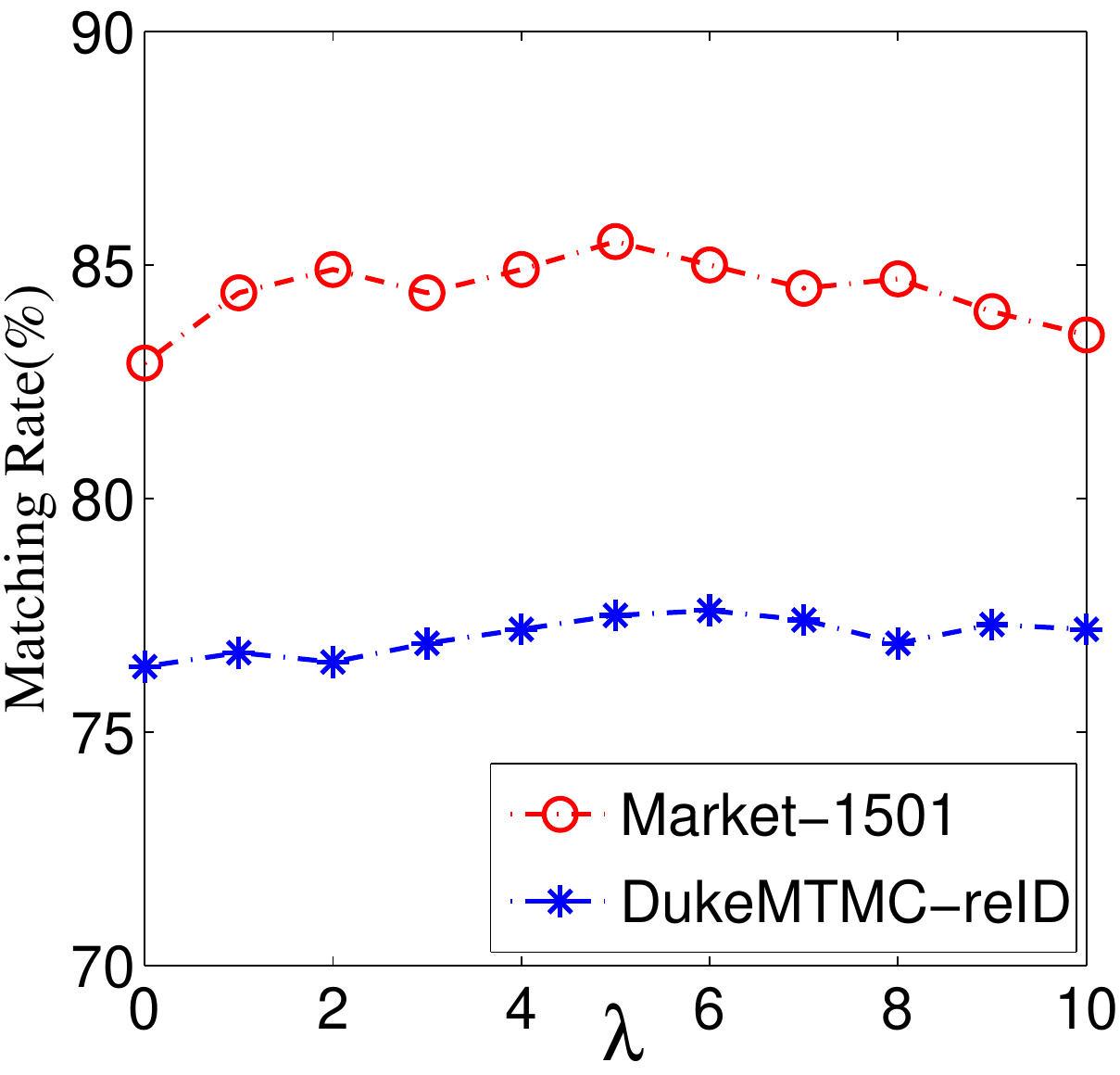}}
%\hspace{1in}
\subfigure[Rank-1]{
\label{fig:subfig:b} %% label for secondsubfigure
\includegraphics[width=1.59in]{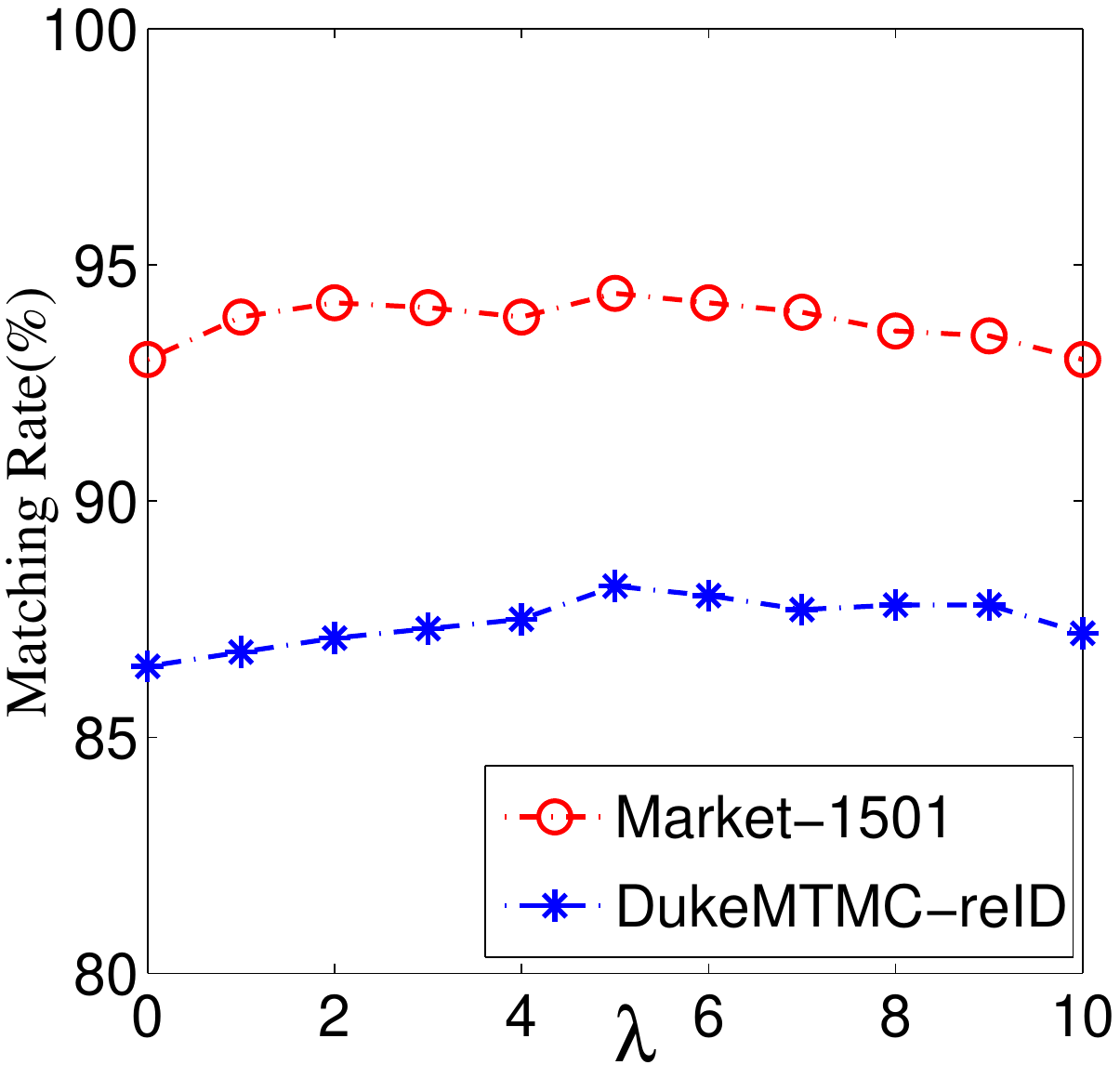}}
\caption{Performance comparison with different parameter $\lambda$ on Market-1501~\cite{Zheng_2015_ICCV} and DukeMTMC-reID~\cite{zheng2017unlabeled}, respectively.}
\label{fig:para} %% label for entire fipgupre
\end{figure}

Here, we evaluate the effectiveness of AttKGCN with different parameter settings.
We first conduct experiments to verify the effect of parameter $\lambda$ (Eq.~(\ref{equ:10})).
Figure~\ref{fig:para} shows performance of our AttKGCN with different parameter $\lambda$.
Here, we can note that AttKGCN results are fairly stable when $\lambda= 4\sim 6$, which
demonstrates the insensitivity of the proposed AttKGCN w.r.t. parameter $\lambda$.
In all experiments, we set $\lambda =5$.

%(2) There are always some range (mostly $\lambda=0.4\sim 0.6$) where AttKGCN obtains
%better results. In our experiments, we fix and set $\lambda=0.5$.
%
%Empirically, the proposed model is insensitive to its parameters.
%
%\textbf{Effects of parameter $\lambda$}. To verify the effects of different values of $\lambda$ in Eq.(10) for balance the contribution of the identification loss and attribute recognition loss.
%As shown in Figure 4, person re-ID classification will playing an increasingly important role as the value of $\lambda$ becomes larger and larger until $\lambda = 5$ that begin to decline on two benchmarks.
%Thus, we set $\lambda = 5$ during our experiment.
%
% When we slightly adjust the parameters, the final tracking results only change a little. Table IV shows the results of the proposed method with different parameters. One can note that, when the parameters are slightly
%changed, our method can maintain good performance, which demonstrates the insensitivity of the proposed method w.r.t. its parameters.
%
%% \textbf{Effects of the depth of GCN}.

We then conduct experiments to verify the effect of  different number of graph convolutional layers in the GCN module of AttKGCN.
Table~\ref{table:3} shows the performance of our AttKGCN  method across different number of convolutional layers.
One can note that AttKGCN can obtain better performance with different numbers of layers, which indicates the insensitivity of
the AttKGCN w.r.t. model depth.
In all experiments, we use a two-layer graph convolutional network in our AttKGCN model. %$\lambda =5$.

% for our performance results.
%In our experiment, we set the output dimensions of each layer as 512, 1024 and 2048 when three-layer GCN model used, respectively.
%And set the output dimensions as 256, 512, 1024 and 2048 when four-layer GCN model used.
%And so on.
%As shown in Table 4, the performance of re-identification decreases with the increase of the number of graph convolution layers.
%The possible reason is our nodes too little.
%When we compute the propagation between nodes, it may be over-smoothing.
%During our experiment, we use two-layer on proposed AttKGCN without otherwise stated.

\begin{table}[]
\caption{Performance comparison with different depths of GCN in our AttKGCN model on two benchmarks.}
\begin{center}
\vspace{-0.35cm}
\begin{tabular}{c|c|c|c|c|c|c}
\hline
\multirow{2}{*}{Layer} & \multicolumn{3}{c|}{Market-1501} & \multicolumn{3}{c}{DukeMTMC-reID} \\ \cline{2-7}
                       & R1        & R5        & mAP      & R1         & R5        & mAP       \\ \hline
2-layer                &\textbf{94.4}      &\textbf{98.0}      &\textbf{85.5}     &\textbf{87.8}       &\textbf{94.4}      &\textbf{77.4}      \\ \hline
3-layer                & 94.0      & 97.9      & 85.1     & 86.8       & 93.9      & 77.1      \\ \hline
4-layer                & 94.0      & 97.7      & 84.8     & 87.6       & 94.3      & 77.0      \\ \hline
5-layer                & 94.3      & 97.7      & 85.1     & 86.4       & 93.6      & 75.2      \\ \hline
\end{tabular}
\end{center}
\label{table:3}
\end{table}
% ablation anaysize
\begin{figure}
\centering
\subfigure[mAP]{
\label{figa} %% label for first subfigure
\includegraphics[width=1.59in]{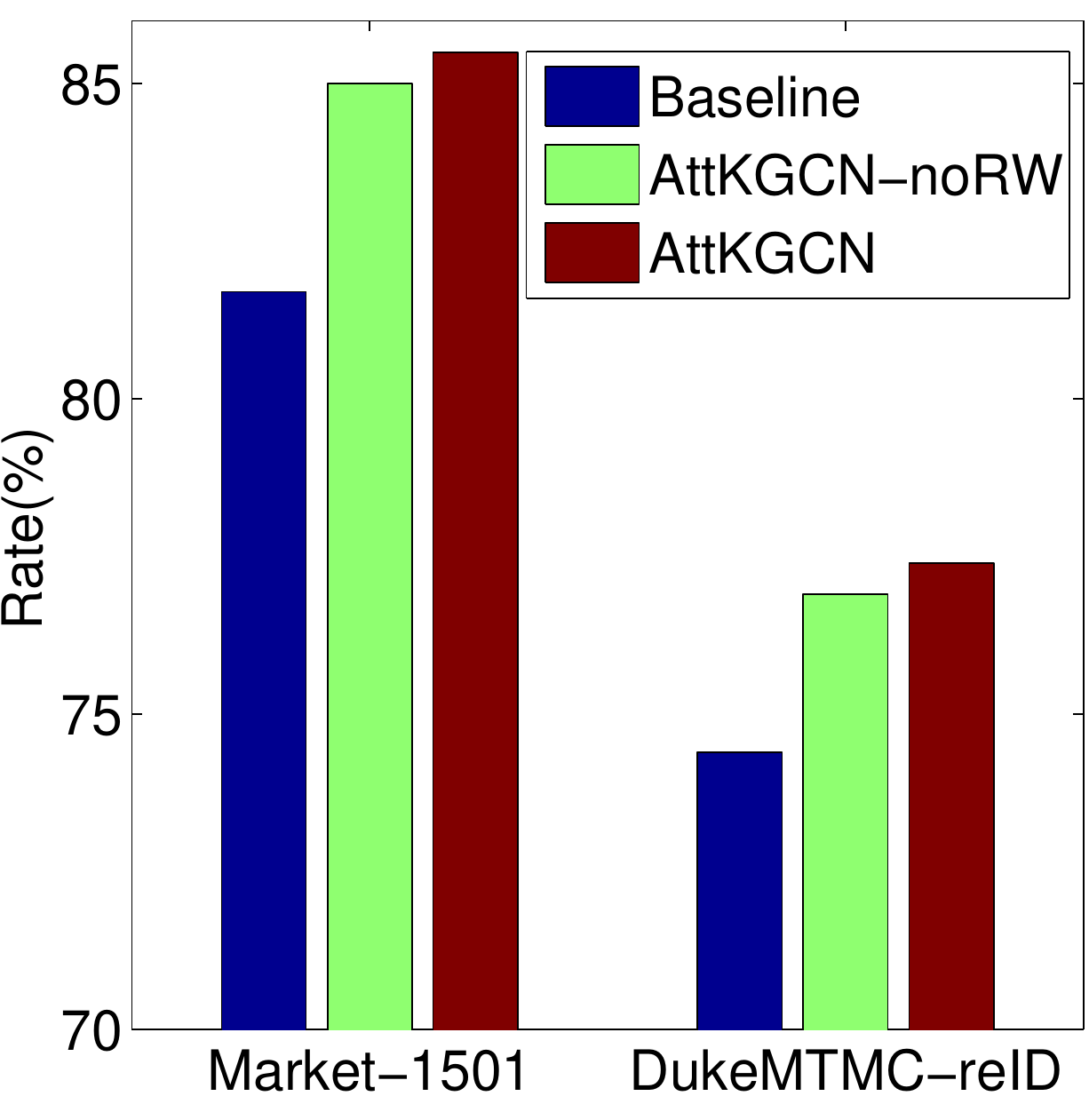}}
%\hspace{1in}
\subfigure[Rank-1]{
\label{fig:subfig:b} %% label for secondsubfigure
\includegraphics[width=1.59in]{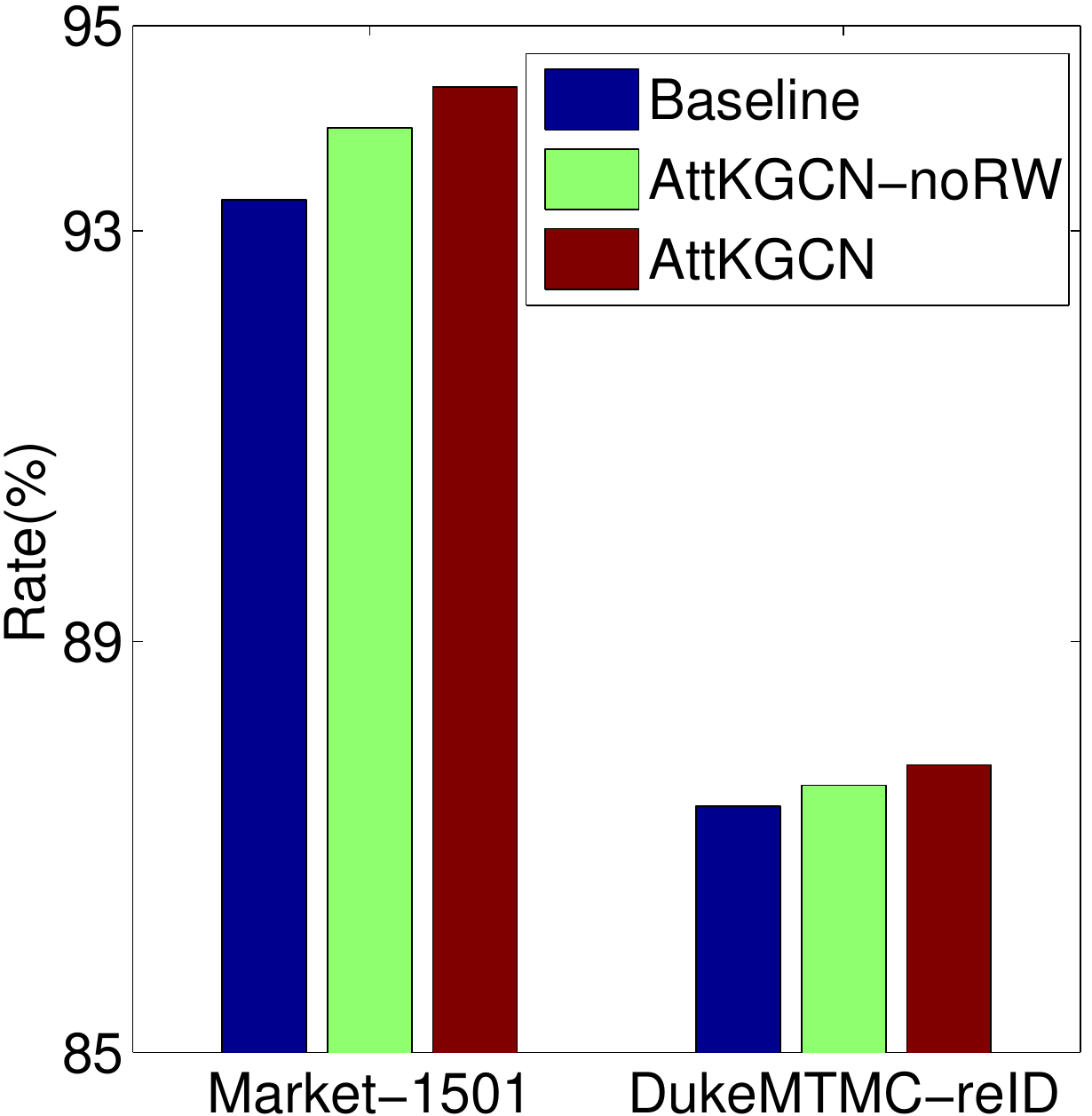}}
\caption{Comparison of Baseline, AttKGCN-noRW and AttKGCN on Market-1501~\cite{Zheng_2015_ICCV} and DukeMTMC-reID~\cite{zheng2017unlabeled}, respectively.}
\label{fig:ablation} %% label for entire fipgure
\end{figure}

\subsection{Ablation Study}
\label{sec:4.5}

To justify the effectiveness of two main components (attribute knowledge graph convolutional module, attribute re-weighting module) in the proposed AttKGCN model,
we conduct ablation experiments on two datasets.
We implement some special variants of our model, i.e., Baseline, AttKGCN-noRW and AttKGCN. % Ours-noT, Ours-noG and Ours-noR.
%1) Ours-noF only conducts graph label propagation and does not use feature of patches our patch weight computation model.
1) Baseline model only uses the visual representation of person image for Re-ID problem and does not exploit any attribute information. % the temporal correlation among frames in our model.
2) AttKGCN-noRW that removes attribute re-weighting module from AttKGCN.
Figure~\ref{fig:ablation} shows the comparison results.
Here, we can note that (1) attribute knowledge graph convolutional module can significantly improve the final Re-ID results, which indicates the
advantage of this architecture on conducting Re-ID tasks.
(2) The proposed attribute re-weighting module is useful to guide the more accurate attribute representation and Re-ID. %matching prediction with one-to-one matching constraints.

%\emph{} PASCAL VOC dataset.

%\emph{Ineffectiveness of  AttKGCN module}.
%From Table 1, we can observe that our results have significant performance improvements over the baseline without AttKGCN.
%For example, we get 43.8$\%$ mAP and 67.9$\%$ rank-1 for Baseline 1 on Market-1501~\cite{Zheng_2015_ICCV}.
%The result of our proposed AttKGCN improves 6.2$\%$ and 5.1$\%$ on mAP and rank-1 compared with Baseline 1, respectively.
%On DukeMTMC-reID~\cite{zheng2017unlabeled}, Baseline 1 has 36.0$\%$ mAP and 55.5$\%$ rank-1.
%Compare with it, our AttKGCN gain 5.1$\%$ and 8.4$\%$ improvements on mAP and rank-1, respectively.
%It verifies that the effectiveness of adding attribute prior information to promote Re-ID task.
%
%\emph{Effects of the Attribute Re-weighting}.
%To verify the effectiveness of our attribute re-weighting (AttRW) module, we implement a variant of AttKGCN, i.e., AttKGCN-noAttRW that removes AttRW from  AttKGCN. %our AttKGCN model named AttKGCN(w/o AttRW).
%We experiment on two benchmarks.
%As shown in Figure 5, you can observe the performance of AttKGCN exceed that of AttKGCN without AttRW model.
%On Market-1501, we improve the accuracy of AttKGCN with AttRW 3.5$\%$ and 0.6$\%$ on mAP and Rank-1, respectively.
%On DukeMTMC-reID, we also improve 0.5$\%$ and 0.2$\%$ on mAP and Rank-1, respectively.
%Experiments adequately proved the effectiveness of our AttRW we add.

\section{Conclusions and Future Works}

This paper proposes a novel Attribute Knowledge Graph Convolutional Network (AttKGCN) model for person Re-ID and attribute recognition.
AttKGCN employs a novel Attribute Knowledge Graph convolutional architecture for attribute representation and learning.
AttKGCN integrates attribute representation and image visual representation together to learn a stronger discriminative person representation for Re-ID.
Experimental results on benchmarks demonstrate that AttKGCN performs obviously better than state-of-the-art attribute-based Re-ID approaches.
Note that, our AttKGCN is not limited to deal with person Re-ID.

In the future, we will adopt AttKGCN to address some other object Re-ID tasks, such as vehicle Re-ID.
Also, we will incorporate some more semantic embedding of attributes (e.g., word embedding) into AttKGCN framework to further enhance the accuracy of attribute recognition

\ifCLASSOPTIONcaptionsoff
  \newpage
\fi



%\bibliographystyle{IEEEtran}
%\begin{thebibliography}{1}


%\end{thebibliography}

\bibliographystyle{IEEEtran}
\bibliography{ref}


\begin{comment}
\begin{IEEEbiography}{Michael Shell}
Biography text here.
\end{IEEEbiography}

\begin{IEEEbiographynophoto}{John Doe}
Biography text here.
\end{IEEEbiographynophoto}

%\newpage

\begin{IEEEbiographynophoto}{Jane Doe}
Biography text here.
\end{IEEEbiographynophoto}
\end{comment}

%\vfill


\end{document}